\documentclass[final]{cvpr}

\usepackage{times}
\usepackage{epsfig}
\usepackage{graphicx}
\usepackage{amsmath}
\usepackage{amssymb}

\usepackage[pagebackref=true,breaklinks=true,colorlinks,bookmarks=false]{hyperref}

\def\cvprPaperID{11271} 
\def\confYear{CVPR 2021}

\begin{document}

\title{Convolutional Spiking Neural Networks \\
for Spatio-Temporal Feature Extraction}

\author{Ali Samadzadeh$^1$, Fatemeh Sadat Tabatabaei Far$^1$, Ali Javadi$^1$,\\ Ahmad Nickabadi$^1$, Morteza Haghir Chehreghani$^2$\\
$^1$Amirkabir University of Technology, Tehran, Iran\\
$^2$Chalmers University of Technology, Gothenburg, Sweden\\
{\tt\small \{a\_samad, tabatabaeifateme, fleissig, nickabadi\}@aut.ac.ir, morteza.chehreghani@chalmers.se}

}

\maketitle

\begin{abstract}
Spiking neural networks (SNNs) can be used in low-power and embedded systems (such as emerging neuromorphic chips) due to their event-based nature. Also, they have the advantage of low computation cost in contrast to conventional artificial neural networks (ANNs), while preserving ANN’s properties. However, temporal coding in layers of convolutional spiking neural networks
and other types of SNNs has yet to be studied. In this paper, we provide insight into spatio-temporal feature extraction of convolutional SNNs in experiments designed to exploit this property. The shallow convolutional SNN outperforms
state-of-the-art spatio-temporal feature extractor methods such as C3D, ConvLstm, and similar
networks. Furthermore, we present a new deep spiking architecture to tackle real-world problems (in particular classification tasks) which achieved superior performance compared to other SNN methods on NMNIST (99.6\%), DVS-CIFAR10 (69.2\%) and DVS-Gesture (96.7\%) and ANN methods on UCF-101 (42.1\%) and HMDB-51 (21.5\%) datasets. It is also worth
noting that the training process is implemented based on variation of spatio-temporal backpropagation explained in the paper.
\end{abstract}

\section{Introduction} \label{section:introduction}
Traditional or Analog Neural Network (referred to ANNs in this paper) has helped AI field reach many paramount goals. ANNs have progressed a lot in many applications. In computer vision, Convolutional Neural Networks have been a significant breakthrough and helped to solve numerous complex tasks. Tasks like large-scale image classification (Imagenet image classification challenge) \cite{touvron2019fixing, xie2020self}, generating real-life images \cite{karras2020analyzing, razavi2019generating, vahdat2020nvae} and many other impressive tasks that were almost impossible to solve without CNNs. Despite their incredible potential, they have some shortcomings; the shortcomings include extreme computation power requirements and lack of memory. Many solutions have been presented to tackle these problems. To solve the problem of extreme computation power usage, \cite{howard2017mobilenets, sandler2018mobilenetv2} have been proposed. Also, to solve memory problems inside each neuron, ConvLSTMs has been the solution so far. However, the mentioned solutions are not perfect. For example, a practical convLSTM has in order of million parameters and will occupy in order of Gigabytes memory which makes it impossible to implement on an embedded vector processing unit. Another example is the MobileNet architecture which still struggles to cope with temporal dimension of data and tackle activity recognition tasks while optimizing accuracy and computational power altogether. There is a better solution called Spiking Neural Networks (SNNs) that solves both problems in a more straightforward manner.

\begin{figure}[t] 
\begin{center}
    \includegraphics[width=0.9\linewidth]{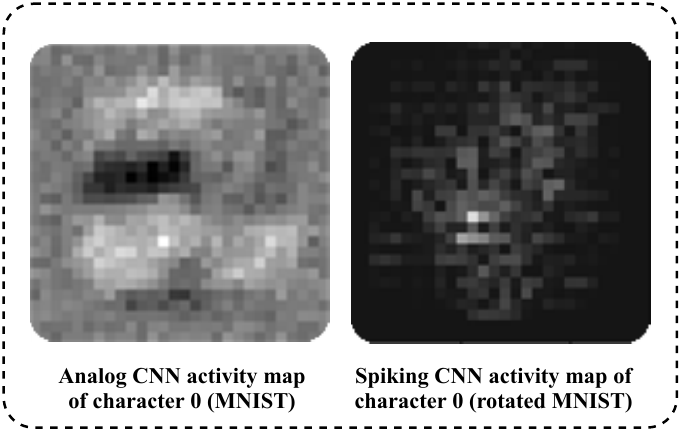}
\end{center}
  \caption{Feature map comparison of SNNs and ANNs. This feature map belong to training on synthetic dataset of rotated character of MNIST (more details about this synthetic dataset in Section \ref{subsection:dataset})}
 \label{fig:frontpage}
 \vskip -0.2in
\end{figure}

Spiking Neural Network encodes data in sequences of spike signals. It may execute more complex cognitive tasks in a way that becomes more similar to the brain cortex processing pattern \cite{allen2009cognitive, kasabov2015spiking, zhang2013spike}. When a neuron's membrane potential reaches a threshold, it is triggered and transmits a spiking signal. To be precise, spikes are binary codes which decay in time (like a electrical capacitor's charge). This binary nature of Spiking Neural Networks makes them efficient in terms of memory consumption and computation cost which leads to lower power consumption (as demonstrated in \cite{wang2020temporal}).

\begin{figure}[t]
\begin{center}
    \includegraphics[width=1\linewidth]{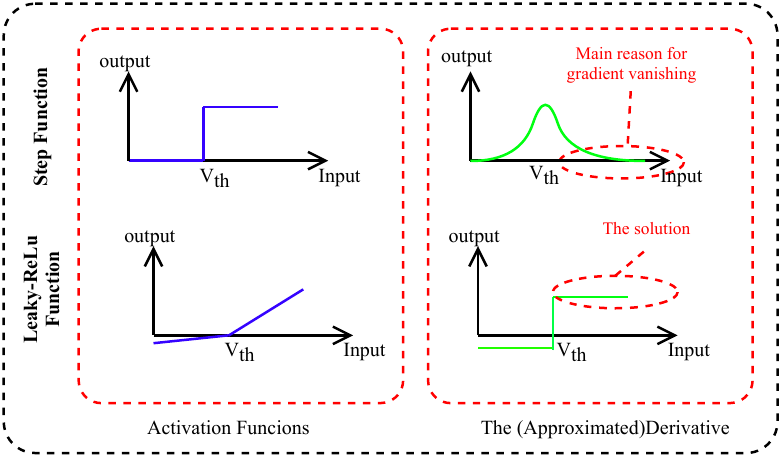}
\end{center} 
  \caption{Gradient vanishing phenomenon happened because of the step function and its approximate derivative in the proposed BPTT methods. Compared to great solution like leaky-relu function.}
  \label{fig:gradientvanish}
\vskip -0.2in
\end{figure}

Despite SNNs' progress, training them is still challenging and an active area of research. There are some great attempts in the literature, namely Spiking Timing Dependent Plasticity (STDP), ANN-SNN conversion and backpropagation though time (BPTT) \cite{mostafa2017supervised, wu2018spatio, wu2019direct}. STDP based methods \cite{ hao2020biologically, kheradpisheh2018stdp, lee2018training, tavanaei2019bp} may have some advantages like the ability to train unsupervised but are limited to shallow networks and cannot train deeper ones required to achieve great performance in complex tasks. Conversely, ANN-SNN conversion methods \cite{ cao2015spiking, diehl2015fast, esser2015backpropagation, han2020rmp, rueckauer2017conversion, sengupta2019going, stromatias2017event} have achieved outstanding results with deep architectures (more than 18 layers). However, these methods as mentioned in \cite{sengupta2019going} assume that SNNs are sampling of ANNs in time. This assumption ignores the temporal properties of SNNs altogether. As some prior works \cite{deng2020rethinking, he2020comparing, wu2018spiking} explained, there are some remarkable temporal properties to the SNNs which cannot be achieved with ANN-SNN conversion based training. BPTT based methods solve this problem by using training through time. These methods usually approximate the derivative of spike functions to be able to use backpropagation \eg \cite{wu2018spatio}. This approximation will be the source of gradient vanishing as demonstrated in \ref{fig:gradientvanish}. These vanishing gradients will cause some issues in training deep architectures (more than 16 layers). Some solutions proposed by \cite{fang2020leaky, wu2019direct} to overcome this problem, however they achieve lower performances than our method.
To solve this problem, we propose a better solution by pre-initializing the weights with initial non-spiking training and using skip connections like the ResNet architecture (more explanation in section \ref{subsection:deepnn}). Furthermore, as \cite{deng2020rethinking} mentioned to demonstrate the spatio-temporal power and efficiency of SNNs, we design a series of test cases (synthetic dataset) to showcase the spatio-temporal property of SNNs compared to ANN ones and make the synthetic dataset a baseline for other recurrent methods to follow (more detail of this synthetic dataset in section \ref{subsection:dataset}). Then, we discuss the spatio-temporal properties of SNNs that differentiate them from conventional recurrent ANNs (see Figure~\ref{fig:frontpage} for instance). Finally, we propose a deep architecture tested on NMNIST \cite{orchard2015converting}, IBM DVS-Gesture \cite{amir2017low}, DVS-CIFAR10 \cite{li2017cifar10}, UCF-101 \cite{soomro2012ucf101} and HMDB-51 \cite{kuehne2011hmdb} datasets and it achieves state-of-the-art performance utilizing the proposed learning method.

\section{Related work} \label{section:related}

Since commercial release of event-cameras in 2008, applications for neural networks with low computation cost have emerged (Spiking Neural Networks to be exact). Interestingly, SNNs can be utilized in real-time applications and harsh environments (i.e. when we have noisy pixel values or against adverserial attacks as \cite{sharmin2019comprehensive} shows in detail). Real-time applications of SNNs consist of but not limited to visual simultaneous localization and mapping or visual odometry (also knows as VSLAM or VO) \cite{kim2016real, kueng2016low, rebecq2016evo}, pose tracking applications \cite{gallego2017event, mueggler2014event} and etc. Furthermore, they are useful in high-speed applications such as object recognition in self-driving cars \cite{wang2020temporal}. Moreover, due to very low power consumption and low latency and lighting condition robustness of event cameras, applications of SNNs can be extended to other vision domains if event-camera's price drops.

Training SNNs is not that easy, although numerous methods are proposed for training them in recent years. Most studied methods focus on converting weights of an ANN model to equivalent SNN \cite{diehl2015fast, esser2015backpropagation, rueckauer2017conversion, stromatias2017event}. For example, \cite{hu2018spiking} and then \cite{sengupta2019going} converted deep residual ANN architecture to their SNN counterparts achieving near state-of-art performance. The ANN-SNN conversion methods may have a very good performance but they have two evident problems: first, they are less robust to noise and adversarial attacks (more details in \cite{sharmin2019comprehensive}) and second, they suppress temporal coding properties of SNNs. To mitigate this problem, the alternative method is to train SNNs directly (to be precise, by using backpropagation through time). As \cite{neftci2019surrogate} explained, the main problem of direct SNN training is non-differentiability of spiking function. \cite{wu2018spatio} overcame this problem by approximating the derivative of threshold function. 

However, \cite{wu2018spatio} and existing methods similar to it have some challenges with going deeper in Spiking Neural Networks. \cite{wu2019direct} tries to recreate batch normalization for SNNs to use its properties and build deeper networks. Nevertheless, the NeuNorm solution (\cite{wu2019direct}) does not have the same properties as batch-norm and is not much of a help in training deeper networks (more than ten layers).


To summarize, we have tackled the problem of deeper SNNs via pre-initialization of weights BPTT proposed in \cite{wu2018spatio} (more details in section \ref{subsection:trainmethod}). We also adapted the method in \cite{sengupta2019going} to create deep residual SNNs and solve complex classification tasks (such as DVS-CIFAR10). 


\section{Spatio-temporal property of SNNs} \label{section:properties}

In this section, LIF model is explained in detail and is compared to recurrent ANN architectures. Moreover, a special synthetic dataset is designed to showcase the critical differences between SNNs and pier ANNs in terms of spatio-temporal feature extraction.

\subsection{Leaky integrate and fire (LIF)} \label{subsection:lif}
One of the successful and excellent and simple modelings of SNNs is leaky integrate and fire (LIF) model. 
The model, from LIF neurons implementation perspective, is defined as follows:
\vskip -0.2in
\begin{align}
U^{(t,n)} &= \alpha(J-O^{(t-1,n)})U^{(t-1,n)} + g(O^{(t,n-1)}) \label{eq:1} \\
O^{(t,n)} &= f(u^{(t,n)})  \label{eq:2}
\end{align} 

In Eq (\eqref{eq:1}), ${J}$ is a matrix of ones and ${g(.)}$ is the layer-wise operation. For linear layers ${g(.)}$ will be defined as:
\begin{align}
g(O^{(t,n-1)})=W^{n} O^{T (t,n-1)} \label{eq:3}
\end{align} 
The term ${(J-O^{(t-1,n)})}$ in the left side of \eqref{eq:1} is for neuron rest and enforcing sparsity in the LIF neurons. 
${f(.)}$ is the activation function which can be interpreted as a threshold function. This function for each node is defined as:
\begin{align} \label{eq:4}
f(U_{i}^{(t,n)})=\begin{cases}
             1  & \text{if } U_{i}^{(t,n)} \ge T \\
             0  & \text{if } U_{i}^{(t,n)} < T
            \end{cases}
\end{align} 

In the equations above, ${n}$ is the layer number, ${t}$ is the time-stamp and ${\alpha}$ is decay factor in \eqref{eq:1}.~Note that the decay factor needs careful tuning and all of the equations (except the activation function) are in the matrix form. 

\begin{figure} 
\begin{center}
\centerline{\includegraphics[width=2.5in, height=3in]{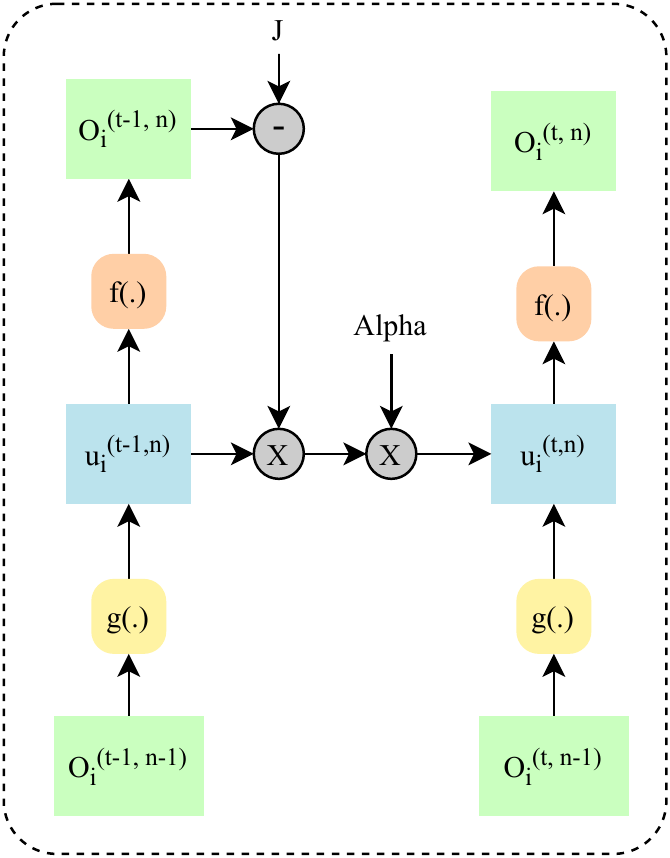}}
\caption{LIF neurons in SNNs, expanded in time and space. $g(.)$ is the analog layer (\eg convolutional layer), $u_i^{(t, n)}$ is the membrane potential of each neuron i inside layer n in time instance of t, $f(.)$ is the activation function which is the step function; therefore, $O_i^(t,n)$ is the output of the neuron i inside layer n in time instance of t, $J$ is the matrix of ones which is used to identify if the neuron is fired to let it stay in rest for the time step and finally ${\alpha}$ is the decay factor which determines how much of the membrane potential (memory in general) should be passed to the next time instance}
\label{fig:lif}
\end{center}
\vskip -0.2in
\end{figure}

The difference between an ANN and SNN neuron is the left side of \eqref{eq:1} and the activation function. The following operation needs to be performed on output of last layer to obtain output of the SNN (assuming rate encoding over an arbitrary time window):
\begin{align}
O^{(t,N)}=\frac{1}{T}\sum_{k=t-T}^{t}{O^{(k,N)}}  \label{eq:5}
\end{align} 
Figure \ref{fig:lif} summarizes the description of LIF model. 
The architectures presented in this paper employ the mentioned LIF neuron model. Furthermore, a synthetic dataset is proposed in this paper to experimentally prove and showcase SNNs' interesting spatio-temporal properties. In the next section, this synthetic dataset is explained in details.

\subsection{Synthetic dataset} \label{subsection:dataset}

\begin{figure}[t] 

\vskip 0.2in
\begin{center}
\centerline{\includegraphics[width=\columnwidth]{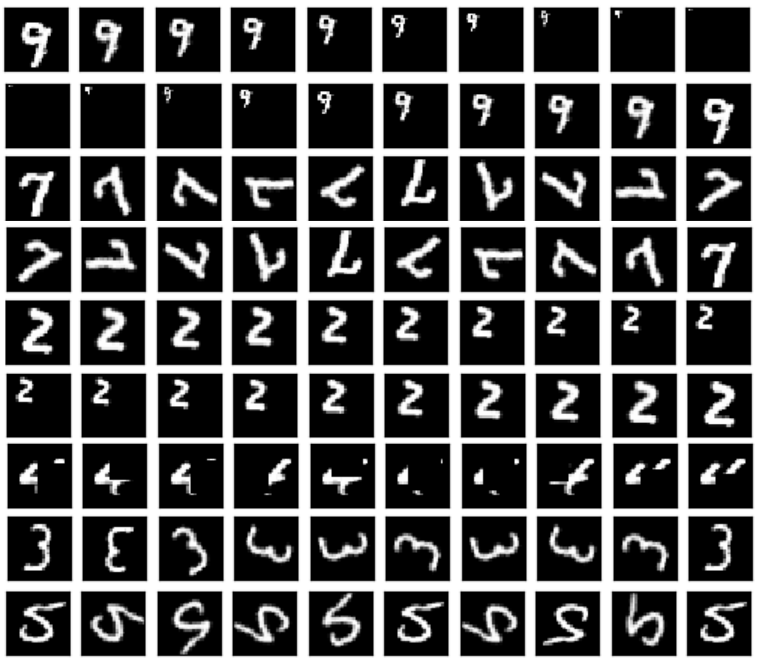}}
\caption{Synthetic dataset designed to challenge spatio-temporal extraction properties. The base images are derived from MNIST dataset. The columns from left to right indicate time which is limited to a window with length of 10. The frames in top two rows belong to sequence1(zoom-in from 0\% to 100\% and zoom-out from 100\% to 0\%), frames in the 3rd and 4th rows belong to sequence2(360 degree clock-wise and counter clock-wise rotations from 0 degree to 360 degree and vice-versa), frames in 5th and 6th rows belong to sequence3 (zoom-in from 50\% to 100\% and zoom-out from 100\% to 50\%), the frames in the 7th row belong to sequence4 (occlusion of the character) and the last two rows in the bottom belong to sequence5(random rotation clock-wise and counter clock-wise).}
\label{fig:mnist}
\end{center}
\vskip -0.2in
\end{figure}

As earlier in the paper mentioned, a synthetic dataset is designed to demonstrate the spatio-temporal feature extraction property of Spiking Neural Networks (see Figure~\ref{fig:mnist}). This synthetic dataset is designed using MNIST numbers and consists of 5 cases, each challenging an aspect of spatio-temporal feature extraction. 

In detail, sequence 1 is a 0 to 100 percent zoom-in and zoom-out act performed in a time window of 10, each one 10 percent larger or smaller (based on zoom-in or zoom-out) than the previous one (see Figure~\ref{fig:mnist}, row 1 and 2, to get a better sense). In this sequence, the network should classify the number and the act which is a zoom-in or zoom-out act; therefore, there are 20 classes to be specified (two zoom-in and zoom-out on ten characters of MNIST dataset). The challenge here is to remember the zoomed-out number (long term memory required). 

Sequence 2 is a regular rotation from 0 to 360 degrees and reverse within the same time window of ten. Each frame is 36 degrees rotated version of its previous frame (clockwise or counterclockwise). In Figure~\ref{fig:mnist}, row 3 and 4, a 7 character from MNIST being rotated inside a time window of 10 is demonstrated.

Sequence 3 is very similar to sequence 1 with the difference in zoom percentage. In this sequence, the zoom-in (and out) is performed from 50 percent (to 50 percent); therefore, basically, the challenge of sequence 1 is less evident here, and the methods should only be able to recognize the difference of almost subtle zooming-in (or zooming-out) inside a time window. 

Sequence 4 is designed to challenge the networks in terms of noise robustness and occlusion happened in many scenarios. In this sequence, the characters of MNIST are masked by a square with dimensions of 14x14 pixels (half the dimensions of the original character). This mask is moved randomly over the characters inside the time window of 10. Therefore, this sequence consists of ten classes of occluded MNIST videos. Using this masking dimension, it is almost impossible for the network to determine the class of characters and reach high accuracy without any memory (because for some characters, like 8 and 3, there is no clear frame that can specify the exact character inside the window). An example of character 4's frames inside the time window is available in Figure~\ref{fig:mnist} row 7.

Lastly, sequence 5 is the most challenging sequence in terms of the human eye (see Figure~\ref{fig:mnist} rows 9 and 10 to get a great sense of the challenge). In this sequence, each character is rotated clockwise or counterclockwise with a random degree between 0 and 360 degrees. This random selection is based on the uniform distribution (therefore, the mean rotation would be 180 degrees). The rotations are also performed in a time window of 10. Similar to sequence 2, this sequence also consists of 20 classes (counterclockwise or clockwise rotated frames of MNIST characters). This sequence challenges the methods that use regular rotation patterns in frames because there are no two equal rotation patterns in the train and test part of this sequence.

In summary, this synthetic dataset can verify most aspects of spatio-temporal feature extraction in a simple way.

\begin{figure*} 
\begin{center}
\includegraphics[width=0.8\textwidth]{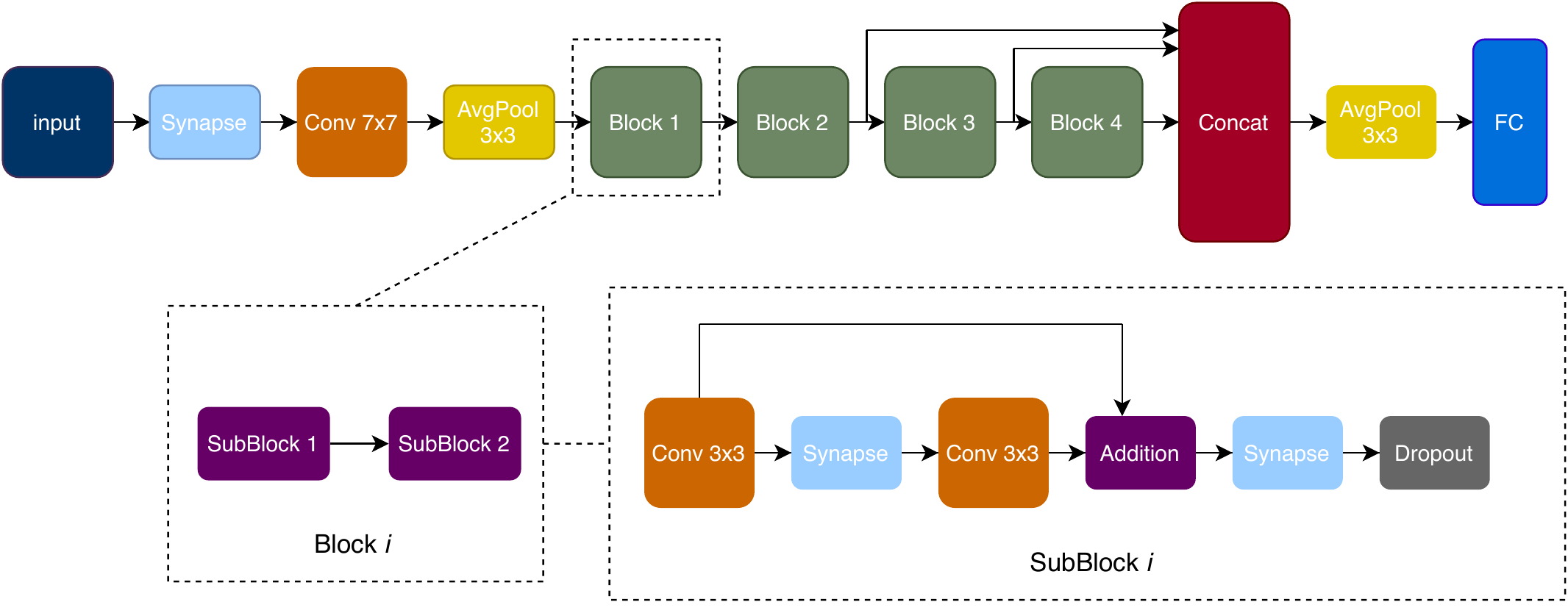}
\end{center}
   \caption{Proposed Deep SNN architecture (STS-ResNet).} 
   \label{fig:architecture}
\end{figure*}

\section{Traning Deep SNNs} \label{section:train}
This section provides a new method to train deep Spiking Neural Network architectures. Then, a new architecture is proposed to prove usablity of the training method that reaches the state of art performance. This architecture, shown in Figure~\ref{fig:architecture}, is inspired by Resnet architecture and the work of \cite{sengupta2019going}.

\subsection{Training method} \label{subsection:trainmethod}
As explained in Section~\ref{section:related}, ANN-SNN conversion methods achieve very high accuracy over very complex datasets like Imagenet, although they do not showcase SNNs' temporal property and don't have such noise robustness as BPTT methods. 
The proposed training method is almost a hybrid of both and has a few differences from BPTT proposed in \cite{wu2018spatio}. To be specific, at first $n$ epochs, the SNN model is altered to be trained similar to an ANN without the problem of gradient vanishing. To do this step, the output activation function of all layers but the output layer is changed to shifted leaky-relu. Also, there is no resting mechanism embedded. So, basically the Eq.~\ref{eq:1} and \ref{eq:4} change into the Eq.~\ref{eq:6} and \ref{eq:7}. In the shifted leaky relu function demonstrated in Eq.~\ref{eq:7}, the $\beta$ hyper-parameter is a small constant near zero. 

\vskip -0.2in
\begin{align} \label{eq:6}
U^{(t,n)} &= \alpha U^{(t-1,n)} + g(O^{(t,n-1)})
\end{align} 
\begin{align} \label{eq:7}
f(U_{i}^{(t,n)})=\begin{cases}
             U_{i}^{(t,n)}  & \text{if } U_{i}^{(t,n)} \ge T \\
             -\beta U_{i}^{(t,n)}  & \text{if } U_{i}^{(t,n)} < T
            \end{cases}
\end{align}

After $n$ epochs, the activation function is switched into the step function and LIF neuron model is used. From this point forward, the outputs are binary as they should be in an SNN and the training procedure is similar to BPTT proposed in \cite{wu2018spatio} and \cite{wu2019direct}. The proposed inital step will make the training much more robust (as depicted in Figure~\ref{fig:trainacc}) and will enable the deep Spiking Neural Networks to reach much higher accuracy; therefore, almost resolving the problem of gradient-vanishing as demonstrated in Figure~\ref{fig:gradientvanish}.
In the following section, a deep SNN architecture is proposed to showcase this training method's efficiency and state-of-art performance.

\subsection{The STS-ResNet architecture} \label{subsection:deepnn}
As mentioned in the previous sections, the principal difficulty of training deep SNNs is gradient-vanishing. Resnet architectures, solve the obstacle of gradient vanishing by utilizing the skip connections. The skip connections increase performance at a drastic rate as well. Inspired by \cite{sengupta2019going}, a Resnet18 like architecture is proposed (see Figure~\ref{fig:architecture}). The main difference of our proposed network to the one in \cite{sengupta2019going} is synapse places, dropouts and additional skip connections and the fact that it can be trained directly. Hence, this architecture is called spatio-temporal Spiking Resnet (STS-ResNet).  
Details of each block are shown in Figure~\ref{fig:architecture}. There are two sub-blocks in each block similar to Resnet18 architecture. In order to force binary outputs after each layer, thresholding activation function namely synapse is applied to the output of each layer. Inside subblocks, there is a dropout to ensure generality and force sparsity. Dropouts are somehow playing the role of batch-normalization for generalization. Average pooling layers do not conflict with the nature of binary outputs in each layer. They are part of the next layer operation; therefore, for example the last average pooling layer is part of the FC layer defined as a new layer (a fact never mentioned before in previous works). 
Another major difference of this architecture to ResNet18 is the block skip connections. They are added from blocks 3 and 4 to the input of average pooling layer. In order to increase performance, instead of conventional summation operation, concatenation operation is used afterward. This concatenation does not happen in Resnet skip connections. 

The STS-ResNet architecture consists of 18 layers, 16 layers in the blocks, 2 convolutional layers and a fully connected layer at the top and bottom of architecture respectively. There is a synapse at the input to allow every kind of inputs (binary, grayscale, color image). It is the first time an 18 layer SNN is trained in space and time domain to classify spatio-temporal tasks. 


\begin{table*}[t] 
\vskip 0.15in
\centering
\begin{center}
\begin{small}

\begin{tabular}{|l|l|c|c|c|c|c|c|} 
\hline
Method & Config & MNIST & Seq1 & Seq2 & Seq3 & Seq4 & Seq5\\
\hline\hline
CNN       & 256-256 &  99\%   & 98.89\% &  98.2\%   & 98.8\%   &  98.27\%   & 20.1\% \\
CNN+LSTM  & 256-256(CNN) + 256(LSTM) & 96.84\% & 67.74\% & \textbf{98.93\%}   & 98.96\%  & 94.88\%    & \textbf{91.2\%} \\
ConvLSTM & 256-256 & 99\%   & \textbf{99.11\%} & 98.9\%    & 30.0\%    &  97.43\%   & 40.7\%   \\
C3D       & 256-256 & 99.03\% & 98.49\% & 98.32\%    & 99.17\%  & 97.73\%    & 64.0\%    \\
\textbf{ConvSNN} & 48-48 & \textbf{99.4\%} &  98.6\%  &  98.4\%   & \textbf{99.36\%}  & \textbf{98.8\%}     & 89.5\% \\
\textbf{STS-ResNet}   & 18 layers (see Sec~\ref{subsection:deepnn}) & \textbf{99.7\%} &  \textbf{99.26\%}  &  \textbf{99.2}\%   & \textbf{99.43\%}  & \textbf{99.1\%} & \textbf{92.7\%}  \\
\hline
\end{tabular}
\end{small}
\end{center}
\caption{Classification accuracy over synthetic dataset explained in Section~\ref{subsection:dataset}}
\label{mnist-table}
\vskip -0.1in
\end{table*}

\section{Experiments} \label{section:experiments}
\subsection{Spatio-temporal feature extraction experiments} \label{subsection:stexperiments}

\begin{figure}[t] 
\begin{center}
    \includegraphics[width=0.8\linewidth]{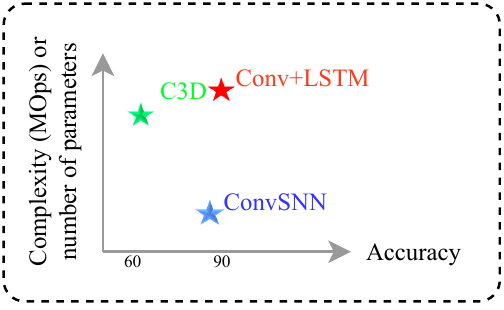}
\end{center}
  \caption{Accuracy Vs complexity (MOps) of tested architectures (CNN, C3D, ConvLSTM, Conv+SLTM, ConvSNN) over the sequence5 of sythetic dataset proposed in Section~\ref{subsection:dataset}. It is evident that the ConvSNN complexity is far less than its competitors while preserving comparable performance.}
\label{fig:accvscomplexity}
\end{figure}

In this section, the goal is to evaluate the performance of top-quality ANN spatio-temporal feature extractors (namely CNN, C3D, ConvLSTM, conv+LSTM) against SNNs. Convolutional Neural Networks are known for their usage in computer vision tasks \cite{tavanaei2019deep}. Moreover, CNN+LSTM \cite{sainath2015convolutional} and C3D network \cite{tran2015learning} are appropriate for modeling spatio-temporal information. ConvLSTM proposed in \cite{xingjian2015convolutional} is also suitable for spatio-temporal feature learning. As mentioned in \cite{srivastava2015unsupervised}, there are some spatio-temporal datasets such as moving MNIST and CIFAR10-DVS to evaluate these methods. Therefore, the synthetic dataset, as explained in Section~\ref{subsection:dataset}, is designed to demonstrate the critical factors of these architectures. For this matter, a shallow network ($<$ 5 layers) of each architecture is used for comparison. Furthermore, to make the comparison more difficult for ourselves, ConvSNN has fewer neurons than it's competitors. Less parameters means less computational power need (as demostrated in Figure~\ref{fig:accvscomplexity}). In the following paragraph, we will explain how we feed the synthetic dataset into the Neural Networks (NNs).

The synthetic dataset is fed into CNN via frame concatenation; therefore, CNN gets a stack of images as input. This type of dataset feeding holds for the C3D too, as we concatenate the frames in 3rd dimension and give it to the C3D network. Time-domain is considered for feeding data into other architectures (ConvLSTM, Conv+LSTM and ConvSNN) and frames are fed into the networks in each time step and then backpropagation through time is performed at the end. The output for these architectures is determined by a voting mechanism which indicates that a neuron with the most amount of firing in the time window will win (also known as the winner-takes-all or fire-rate-based output). For example, considering time window with length of ten, each frame is given to the Convolutional SNN. At the end of each time window the neuron with most firing rate will be the winner. In the following paragraph the essence of memory for NNs is explained.

Due to the structure of synthetic dataset, this frame by frame feeding though time clearly requires the networks to have some sort of memory. To explain this memory need further, take the example of seq2. If the NN has no memory, feeding this sequence frame by frame will cause it to not recognize clockwise and counterclockwise movement. Since it has received the same frames with different orders at the end of the time window. Fire-rate-based output mechanism will not make an acceptable prediction for two mentioned frame sequences. In the following, we will discuss about training methods and evaluation criteria.


Training method used for the ConvSNN is the simple spatio-temporal backpropagation presented in \cite{wu2018spatio}. Training algorithm for other methods were conventional BP and BPTT. The training is performed for at least 100 epochs and the maximum performance on the test set is recorder for each architecture. For the purpose of evaluation, each NN is tested 10 times and the average top-1 accuracy is reported in Table~\ref{mnist-table}. 

As discussed in Section~\ref{subsection:dataset}, the general property of spatio-temporal feature extraction is examined with sequences 2 and 3. All architectures show promising results on those two sequences (Table~\ref{mnist-table}). Then, long-term preservation of data in memory is investigated using seq1 as explained in Section~\ref{subsection:dataset}. Typical LSTM layer (lacking cut connections) does not have this property and therefore the accuracy will drop as shown in Table~\ref{mnist-table}. Moreover, the robustness of architectures to noise and their ability to extract spatial features through time is examined via seq4. The results prove SNN superiority in this aspect over all other networks. The CNN+LSTM performance is considerably worse than the others. The output of CNN is given to the LSTM layer and it does not contain all properties of spatial features; therefore, making it strenuous for the LSTM layer to determine the occluded character.
Finally, testing with seq5, the random spatio-temporal features generated in time challenges the non-repeatable temporal feature extraction qualities of the mentioned NNs. This sequence highlights SNNs great ability in classifying stochastic non-repeating time patterns over the rivals. Results in Table~\ref{mnist-table} show that only CNN+LSTM can outperform SNN in this sequence with little margin, other NNs drastically fail to achieve comparable accuracy, specially the CNN architecture. Although, the conv+LSTM network has too many layers compared to SNN and Figure~\ref{fig:accvscomplexity} can show the difference.

\begin{figure}[t]
\begin{center}
\centerline{\includegraphics[width=0.9\columnwidth]{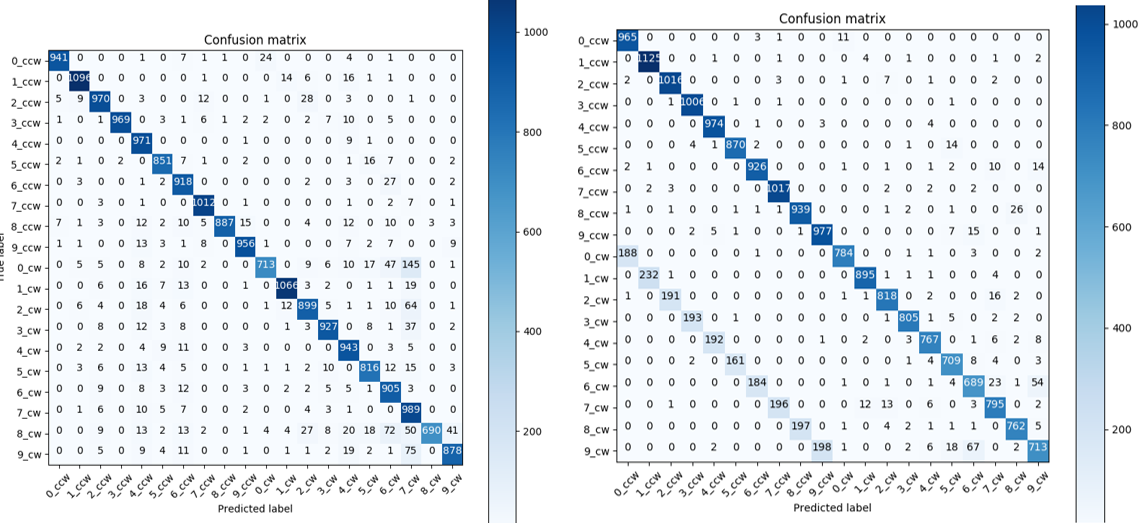}}
\vspace{0.05in}
\caption{Confusion matrix over seq5, comparing result of CNN+LSTM and ConvSNN. The left image shows performance of Conv+LSTM model.}
 \label{fig:conf_a}
\end{center}
\vskip -0.2in
\end{figure}

\begin{figure}[t] 
\begin{center}
\centerline{\includegraphics[width=\columnwidth]{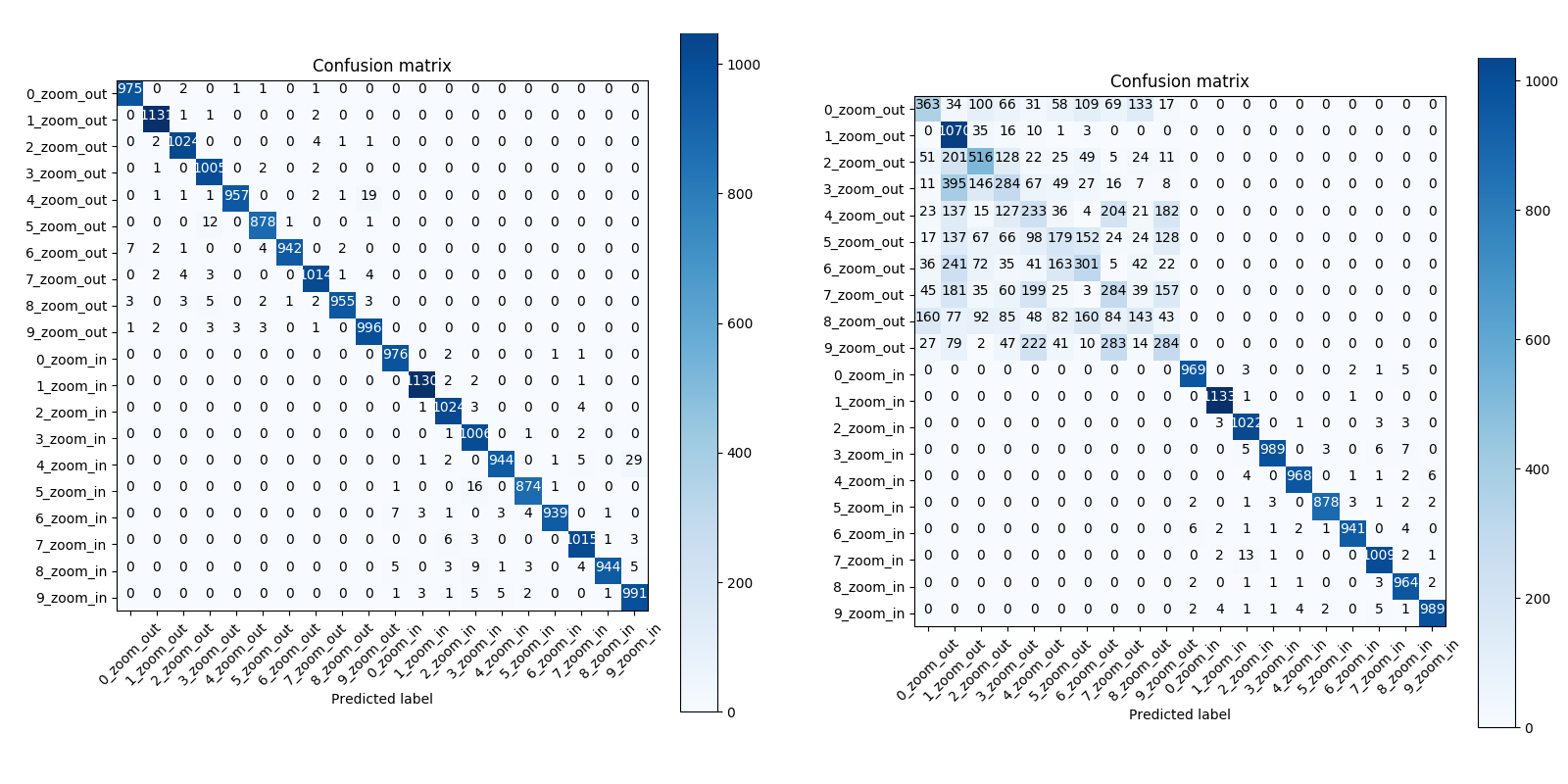}}
\vspace*{-\baselineskip}
\vspace{0.05in}
\caption{Confusion matrix over seq1, comparing result of CNN+LSTM and ConvSNN. The left image shows performance of ConvSNN model.}
\label{fig:conf_b}
\end{center}
\vskip -0.2in
\end{figure}

To further evaluate the amazing achievement of ConvSNN. The confusion matrices were illustrated for sequences 1 and 5 in Figure~\ref{fig:conf_a} and \ref{fig:conf_b}. Figure~\ref{fig:conf_a} illustrates imperfection of convSNNs in extracting random temporal properties (the problem of distinguishing between clockwise and counterclockwise rotation in seq5). Figure~\ref{fig:conf_b} shows long-term memory preservation issue in conventional LSTM layers. Moreover, feature maps of ConvSNN and Conv+LSTM is shown in Figure~\ref{fig:frontpage} to show the vast difference of ConvSNNs and ConvANNs in a short example. As Figure~\ref{fig:frontpage} depicts, SNN architecture tries to extract both spatial and temporal property inside the feature map whereas ANN  only extracts the spatial property of MNIST characters.

\begin{figure}[t] 
\begin{center}
    \includegraphics[width=0.9\linewidth]{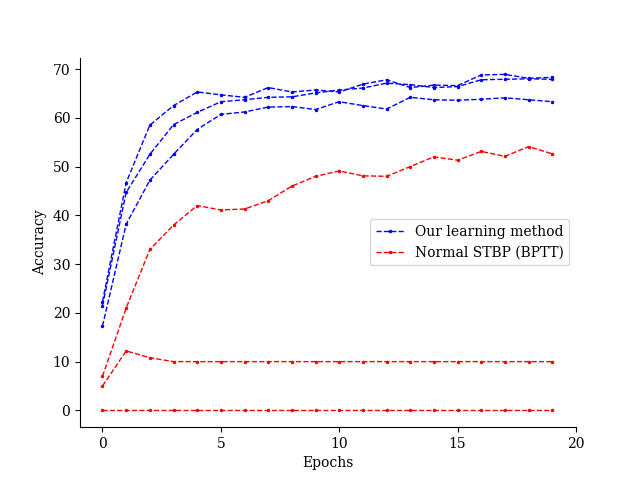}
\end{center}
  \caption{Training accuracy per epoch plot. This plot shows the robustness of performance using our learning method over multiple tries on the DVS-CIFAR10. Three random tries attempted to train the STS-ResNet for each training method. As for our training method, first five epochs where pre-training and the rest is almost equal to STBP.}
  \label{fig:trainacc}
\end{figure}

The mentioned results prove claim of the paper (SNNs are suitable for spatio-temporal feature extraction specially when featuers aren't in a regular time or space pattern or they are noisy). In order to make the SNNs more suitable for complex conditions, we proposed STS-Resnet architecture as mentioned in Section~\ref{subsection:deepnn}. The results of this architecture is reported in Table~\ref{mnist-table} along other methods. The performance of this architecture over more challenging datasets is explained in the next subsection.

\subsection{Experiments with Deep residual SNN} \label{subsection:nnexperiments}

In order to showcase the advantages of our learning method and designed architecture over the prior SNNs and ANNs, we chose NMIST and DVS-Gesture datasets to exhibit the performance of our learning method and architecture over non-complex neuromorphic image classification task. We have compared STS-ResNet with other state-of-the-art (SOTA) SNN methods. Table~\ref{dvs-table} depicts the performance comparison. Furthermore, we chose CIFAR10-DVS dataset to depict performance of this architecture in complex neuromorphic task against the SOTA SNN methods. Finally, we have evaluated the ResNet18 (traditional ANN) and STS-ResNet architecture on complex activity recognition of UCF-101 and HMDB-51 datasets to truly challenge the STS-ResNet architecture against its ANN counterpart.

The output of the networks is calculated based on the fire-rate-based method and the inputs are given to networks in raw form (neuromorphic datasets are fed into NNs with event stacking through time). The evaluation of the networks is based on top-1 accuracy calculated over 10 runs of training for at least 100 epochs.

Table~\ref{dvs-table} elucidates the significant improvement of STS-ResNet over the previous outstanding methods. The only downside of STS-ResNet is being inferior to Resnet18 over the UCF-101 dataset. The reason for this mediocre performance is the extreme importance of spatial features in UCF-101. It is evident from this observation that STS-ResNet, despite being trained by our method, is still inferior to the ANN counterpart; therefore, the training technique is not perfect. On the other hand, the results on HMDB-51 is promising, showing great performance over this extremely challenging dataset. The HMDB-51 dataset has more temporal features embedded than UCF-101 dataset. The temporal features makes the dataset more suitable for SNNs, since they have neuron memories to accommodate for temporal feature extraction. It is worth noting that it is the first time that a SNN architecture is tested on complex datasets like UCF-10 and HMDB-51.

Furthermore, our architecture needs much less kernels than \cite{wu2019direct} and is much more memory efficient. To be precise, our model concentrates on making the network deep, whereas \cite{wu2019direct} tries to increase number of kernels per layer. In order to see the exact parameters, refer to implementation details in the next section.

\begin{table}[t] 
\centering
\begin{center}
\begin{small}
\begin{tabular}{|l|l|c|} 
\hline
Model & Dataset & Accuracy\\
\hline\hline
SPA \cite{liu2020effective}    & NMIST & 96.3\% \\
SpiNNaker \cite{patino2020event}    & NMIST & 98.5\% \\
STBP tested in \cite{patino2020event}  & NMIST & 97.92\% \\
Natural ConvSNN \cite{deng2020rethinking}      &  NMNIST & 99.42\% \\ 
NeuNorm \cite{wu2019direct}      &  NMNIST & 99.53\% \\ 
\textbf{STS-ResNet}    & NMNIST & \textbf{99.6}\% \\
\hline
CNN \cite{amir2017low}      & DVS-Gesture & 94.59\% \\ 
pointnet++ \cite{wang2019space} & DVS-Gesture & 95.32\% \\ 
\textbf{STS-ResNet}    & DVS-Gesture & \textbf{96.7}\% \\
\hline
HAT \cite{sironi2018hats}    & DVS-CIFAR10 & 52.4\% \\
Natural ConvSNN \cite{deng2020rethinking}      &  DVS-CIFAR10 & 60.3\% \\ 
NeuNorm \cite{wu2019direct}      &  DVS-CIFAR10 & 60.5\% \\ 
SPA \cite{liu2020effective}    & DVS-CIFAR10 & 67.8\% \\
\textbf{STS-ResNet}    & DVS-CIFAR10 & \textbf{69.2}\% \\
\hline
ResNet18 (ANN+scratch) \cite{hara2018can}   & UCF-101 & \textbf{42.4}\% \\
\textbf{STS-ResNet}    & UCF-101 & 42.1\% \\
\hline
ResNet18 (ANN+scratch) \cite{hara2018can}    & HMDB-51 & 17.1\% \\
\textbf{STS-ResNet}    & HMDB-51 & \textbf{21.5}\% \\
\hline
\end{tabular}
\end{small}
\end{center}
\caption{Classification accuracy of proposed deep network versus SOTA on NMNIST, DVS-Gesture, DVS-CIFAR10, UCF-101 and HMDB-51 }
\label{dvs-table}
\end{table}

\subsection{Implementation details} \label{subsection:implementation}
In this subsection, we provide the experiment details about networks architectures and parameters.
Basic implementation details for the test cases are as follows:
Frame window size for all networks was set to 10. Learning rate for all network architectures was set to 1e-3 (except ConvSNN which was set to 5e-4). All architectures were trained enough to reach maximum accuracy (more than 100 epochs). These experiments were performed at least 10 times and the mean value for them is reported in the tables. For training networks, Adam optimizer and least mean square error was used (except for ConvLSTM network where Binary cross-entropy was used). Batch sizes were 100 except for ConvSNN which was 20.

As for the ConvSNN specific parameters, threshold value was set to 0.5.~This value is extremely important and slight change in it will result in better or worse results.~${\alpha}$ or decay factor is set to 0.5.~Increasing this value will result in better preservation of memory and vulnerability to noise. Also, resting mechanism was disabled. Resting mechanism maintains more sparisity in spike patterns but results in accuracy drop. Derivative of Dirac function was aproximated with Gaussian function $\mathcal{N}(Threshold, \frac{1}{6})$ (Rect function is better approximation in terms of performance but it will learn harder and the mean accuracy for several runs will drop dramatically). The initial pre-training was performed for almost 50 epochs and then the BPTT presented in \cite{wu2018spatio} was used. The comparison of our learning method and previos BPTT based method is depicted in figure~\ref{fig:trainacc}

Parameters of the proposed STS-ResNet architecture to tackle CIFAR10-DVS are as follows:
Frame window length and learning rate was set to 10 and 5e-4 respectively and training was perfomed for more than 50 epochs and more than 5 runs as before. The optimizer was SGD with momentum of 0.9. Chosen loss function was binary cross entropy. Batch size was 10 and 1000 events were concatanated per frame. Other SNN specific parameters(threshold, resting mechanism, deravative approximate function) were as before, except for decay factor which was set to 0.8 (in CIFAR10-DVS dataset memory is more important than noise robustness).

All of the experiments were performed on a system with Intel Core i5-6500, NVIDIA GTX 1080 and NVIDIA GTX 1080ti with 32 GB RAM and SSD storage.

\section{Conclusion} \label{section:conclusion}
This paper demonstrated the potentials of SNNs in terms of spatio-temporal feature extraction. Particularly, their capacity to extract randomly distributed features in the time and space domain. This claim was supported by experiments with a special type of synthetic dataset designed for the matter. Furthermore, a modified training method is proposed to enable training of deeper networks. Then, a new deep SNN architecture was proposed to showcase its performance gain over multiple known datasets. The introduced SNN architecture was tested on challenging datasets incuding CIFAR10-DVS, NMNIST and DVS-Gesture to depict it's gain over previous architectures.

Regarding the results, a shallow SNN outperformes shallow ANNs over extreme conditions (synthetic dataset), and surpassed SNNs over the typical event-based dataset (CIFAR10-DVS). Moreover, SNNs have much lower memory consumption (with the assumption of binary connections) and computation cost which refers to less overall hardware power consumption. Also, in some situations, SNNs with few number of neurons can achieve what oversized ANNs can barely achieve. 

The remaining problem to be solved is adaptation of batch normalization properties to SNNs. These properties are required to have very deep SNNs \eg 101 layers. Also, there should be a better solution for BP other than approximating the derivative of activation function; the approximate functions are the primary cause of gradient-vanishing. Another step in the journey of analyzing these networks might be an analysis of other types of SNNs such as GANs. 

To sum it all, this work renders the advantages of SNNs transparent and proposes a new deep architecture with some solutions to train other deeper SNNs.

{\small
\bibliographystyle{ieee_fullname}
\bibliography{egbib}
}

\end{document}